\documentclass[10pt, conference, compsocconf]{IEEEtran}

\usepackage{latexsym}
\usepackage{amsfonts}
\usepackage{amssymb}
\usepackage{amsmath}
\usepackage{color}
\usepackage[ruled,vlined]{algorithm2e}

\usepackage{tikz}
\usetikzlibrary{arrows}

\usepackage{graphicx} 
\graphicspath{{./}}
\newcommand{\todo}[1]{{}}

\newcommand{\bigO}{\mathcal{O}}

\newcommand{\T}{\mathcal{T}}

\newcommand{\V}{\mathcal{V}}

\mathchardef\mhyphen="2D
\newcommand{\gengap}{gen\mhyphen gap}
\newcommand{\maxgap}{max\mhyphen gap}

\newcommand{\maxspan}{max\mhyphen span}

\usepackage{makeidx}  % allows for indexgeneration
\begin{document}

\title{A SAT model to mine flexible sequences in transactional datasets\thanks{Thanks to G. Audemard and L. Simon for their help with Glucose source code, T. Guns and S. Nijssen for insightful discussions.}}

% Pour Sylvain: ceci est un commentaire ! 

\author{\IEEEauthorblockN{Remi Coletta}
\IEEEauthorblockA{LIRMM\\
University Montpellier 2\\
Montpellier, France\\
coletta@lirmm.fr}
\and
\IEEEauthorblockN{Benjamin Negrevergne}
\IEEEauthorblockA{Inria\\
Centre Rennes, France\\
benjamin.negrevergne@inria.fr}
}
\maketitle
%
%%
%\frontmatter          % for the preliminaries
%%
%\pagestyle{headings}  % switches on printing of running heads
%\addtocmark{A SAT model to mine flexible sequences in transactional datasets
%} % additional mark in the TOC
%%
%
%\mainmatter              % start of the contributions
%%
%\title{A SAT model to mine flexible sequences in transactional datasets\thanks{Thanks to G. Audemard and L. Simon for their help with Glucose source code, T. Guns and S. Nijssen for insightful discussions.}}
%%
%\titlerunning{A SAT model to mine flexible sequences in transactional datasets}  % abbreviated title (for running head)
%%                                     also used for the TOC unless
%%                                     \toctitle is used
%%
%\author{Remi Coletta\inst{1} \and Benjamin Negrevergne\inst{2} }
%%
%\authorrunning{Remi Coletta and Benjamin Negrevergne} % abbreviated author list (for running head)
%%
%%%%% list of authors for the TOC (use if author list has to be modified)
%\tocauthor{Remi Coletta and Benjamin Negrevergne}
%%
%\institute{
% LIRMM, UMR5506 University Montpellier 2 - CNRS, Montpellier, France\\
%\email{coletta@lirmm.fr}\\
%% WWW home page: \texttt{http://www.lirmm.fr/\homedir\{coletta,vismara\}/}
%\and
%DTAI , KU Leuven, Belgium\\
%\email{benjamin.negrevergne@cs.kuleuven.be}\\
%}
%
%\maketitle              % typeset the title of the contribution
%
\begin{abstract}
Traditional pattern mining algorithms generally suffer from a  lack of flexibility. 
%of traditional pattern mining algorithms is now a well-known issue. 
  In this paper, we propose a SAT formulation of the
  problem to successfully mine frequent flexible sequences occurring in transactional datasets.
  Our SAT-based approach can easily be extended with extra constraints to
  address a broad range of pattern mining applications. To
  demonstrate this claim, we formulate and add several  constraints, such as gap and span
constraints,  to our model in order to extract more specific patterns. We also 
  use interactive solving to perform important derived tasks,
  such as closed pattern mining or maximal pattern mining.  
  Finally, we prove the practical feasibility of our SAT model by running 
  experiments on two real datasets.
%\keywords{}
\end{abstract}

\section{Introduction}

Pattern mining, now both an important research topic and a broadly-used
analysis tool, aims at extracting recurring patterns in large datasets.
Since seminal work by Agrawal and Srikant on frequent itemset mining \cite{agrawal1994fast},
a large number of new pattern mining tasks have been proposed to extract more
{\em structured} patterns, such as sequence, tree or graph patterns.
While structured patterns are usually more informative, they are also harder to extract.%replac eharder = more diffcult ?  

\begin{figure}
  \centering
  \begin{large}
      \begin{tikzpicture}[scale=0.7, transform shape]
        \tikzstyle{seq_item}=[draw, fill=blue!20, shape=rectangle, minimum size=1cm];
        \tikzstyle{seq_item_static}=[draw, fill=white, shape=rectangle, minimum size=1cm];
        \tikzstyle{match}=[out=270, in =90, thick, -latex, color=red!80!black];
            \newcounter{x}

        \node at (0, 0) {$S$}; 
        \setcounter{x}{1}
        \foreach \i in {A, B}{
          \node[seq_item] at (\arabic{x}, 0) (p\arabic{x}) {\textbf{\i}};
          \stepcounter{x}
        }

        \node at (0, -2) {$T_1$}; 
        \setcounter{x}{1}
        \foreach \i in {B, A, C, B}{
          \node[seq_item_static] at (\arabic{x}, -2) (t1\arabic{x}) {\textbf{\i}};
          \stepcounter{x}
        }

        \node at (0, -3.1) {$T_2$}; 
        \setcounter{x}{1}
        \foreach \i in {A, C, C, B}{
          \node[seq_item_static] at (\arabic{x}, -3.2) (t2\arabic{x}) {\textbf{\i}};
          \stepcounter{x}
        }

          % \draw (p1.265) edge[thick,-latex, color=blue!80!black] (t11.95);
          % \draw (p2) edge[thick,-latex, color=blue!80!black] (t12);

        \draw (p1) edge[match] (t12);
        \draw (p2) edge[match] (t14);

        \draw (p1) edge[match, dashed] (t21);
        \draw (p2) edge[match, dashed] (t24);

        % \node[seq_item] (a) {A} ;space
        \end{tikzpicture}
        \end{large}
\caption{Sequence $S = [A, B]$ is embedded in transaction $T_1$ (solid arrows) and transaction $T_2$ (dashed arrows).}
\label{fig:embedding}
\end{figure}
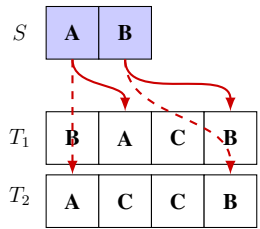

The sequence mining problem is a variation of the itemset mining problem,
in which both transactions and patterns are ordered. Given a set of ordered transactions,
the goal is to find all the sequences that are embedded in more than a given
number of transactions. The complexity of the task stems from the
fact that we allow sequences to be embedded in the transactions with
variable-length gaps. See Figure \ref{fig:embedding} for an
illustration.

%syl: a given number of transactions.

This problem has been given multiple names in the literature: depending on the application,
it has been called {\em embedded subsequent mining}, {\em flexible motifs mining},
or {\em serial episode mining}. Nevertheless, % a garder ? 
it particularly affects bio-medical
applications because of the ubiquity of sequential (DNA or protein) datasets in this domain
(e.g.,\cite{ye2007efficient}). It also impacts text analysis,
where sentences can be seen as sequences of words (e.g.,\cite{DBLP:conf/kdd/TattiV12}).

% In the litterature, there exist many variation of these problems:
% customers sequences are used for market basket analysis, closed

Sequence mining, as well as other pattern mining tasks, was first
developped by writing algorithms in classical imperative
languages such as C or C++. In order to handle large datasets, most
of these algorithms were optimized for specific mining applications.
As a result, they are unable to capture small variations. This is a major drawback
for users, whose needs may substantially vary from one application to the other.
For example, text analysts generally have grammar constraints, whereas biologists usually require very specific, per-character constraints.

To solve this problem, Guns et al. in \cite{DeReadt08} have recently
proposed using CP modeling techniques and CP solvers to address
various itemset mining tasks. In contrast with imperative programs,
CP models can easily be extended to efficiently perform a variety of specific mining tasks,
while remaining more or less performant on the basic mining task. %a revoir 
However, to date, most of the published
works have focused on modeling and solving set-based structures, which are
natively supported by solvers or easy to encode as boolean
vectors. 
%Faire suater phrase suivante ? 
Nevertheless, the question of the usability of ...  needs to be further examined. %mieux 
Whether the same solvers can be used to mine more structured
patterns, such as sequences, trees or graphs using generic solvers,
remains an open question. %remi à revoir . 

\paragraph{Contribution of this paper}
In this paper, we demonstrate that the sequence mining problem 
 can be expressed as a purely boolean SAT formula, and
solved using off-the-shelf SAT solvers. 

To extend the basic problem and address more specific needs,
we formulate several constraints commonly used in
sequence mining. We show how these additional constraints can be easily incorporated
into our SAT formula and without previous SAT knowledge.
This is a critical benefit for users, who can now
arbitrarily combine sets of constraints without concern for the
soundness of their results. In addition, our experiments on real
datasets establish that, contrary to specialized
algorithms, SAT solvers take advantage of these new constraints to
further prune the search space and increase efficiency.

Lastly, we demonstrate that, in addition
to enumerating {\em all} frequent sequences, our model also allows us to perform
important derived mining tasks such as the {\em closed pattern}
and the {\em maximal pattern} mining tasks using interactive
solving. These two tasks aim to reduce redundancy among
result sets and have been extensively studied in the pattern
mining community. 

%   the model can be easily
% extended to address other derived problems such as the problem of
% mining closed sequences or the problem of mining sequences with gap constraints. This
% represent a major benefit over specialized algorithms such as
% BIDE\cite{wang2004bide}, which cannot be adapted to other tasks
% without important implementation efforts.
%\paragraph{}

% We propose a solution to extract {\em closed} and {\em maximal}
% sequences with iterative solving, which are two well-know derived
% problems that cannot be solved with a single SAT formula.

% We finally run experiments using the Glucose SAT solver
% \cite{} on real datasets. This experiments demonstrate
% the feasibility of mining sequences using SAT solvers
% despite being a notoriously complex problem. 

The remainder of this paper is organized as follows: we formally state the problem
in Section~\ref{sec:freq-sequ-mining}.  We then present our SAT model to mine flexible sequences in
Section~\ref{sec:model} and formulate several user constraints in
Section~\ref{sec:solv-deriv-sequ}. We propose our solution to mine
maximal and closed patterns in Section~\ref{sec:maxim-clos-patt}, and
%present several practical optimizations in Section~\ref{sec:exploting} and 
present our experiments in Section~\ref{sec:experiments}.
Finally, we discuss related work in Section~\ref{sec:related-work} and offer our conclusions 
in Section~\ref{sec:conclu}.

\section{Frequent sequence mining: formal problem statement}
\label{sec:freq-sequ-mining}

Given:
\begin{enumerate}
\item a vocabulary $V : \{v_0, \ldots,  v_{|V|}\}$ %remi with $v_0 = \epsilon$, ça intervient seulement dans le modèle
\item a set of transactions $\mathcal T = \{T_1, \ldots, T_{|\T|}\}$ where each transaction $T_i$ is a sequence of characters in $V$;
we use the projection notation $T_i[j]$ to denote the $j^{th}$ character of the transaction $T_i$; and % And we define $l$ as the length 
% of the longest transaction: $l=max_{T_i \in \T} |T_i|$
\item a minimum frequency threshold $minsup$,
\end{enumerate}
%syl: exemples of "enumerate"?
\noindent the sequence mining problem consists in
%can be stated as follows. 
 enumerating all the sequences that are {\em embedded} in at
least $minsup$ transactions.

In the standard sequence mining problem,
we say that a sequence $S$ of $n$ characters is embedded in a transaction $T$ if there
exist $j_1, \ldots, j_n$ integers such that:
$$j_1 < \ldots < j_n,  \forall i \in 1,\ldots,n ~s.t.~ S[i] = T[j_i]$$

The list of positions $e = (j_1, \ldots, j_n)$ is called an {\em embedding}.
To express that $e$ is an embedding of $S$ in $T$, we note $S \sqsubseteq_e T$. 
We also note $S \sqsubseteq T$ if there exists at least
one embedding $e$ such that $S \sqsubseteq_e T$.\\

We now define the {\em coverage} of a sequence $S$ as
the set of transactions for which there exists an embedding of $S$: 
$$cover(S, \mathcal T) = \{T_i \in \T : S \sqsubseteq
T_i\}$$
%Syl: perdu le saut à la ligne de define. Should we not write coverage of a sequence instead?
Using these notations, the problem of mining all frequent sequences in
a set of transactions can be formalized as follows:\\
Given a vocabulary $V$, a set of transactions $\T$ and a minimum
threshold $minsup$, enumerate all sequences $S$ such that:
$$|cover(S, \mathcal T)| \ge minsup$$

% The general problem that we address in this paper is
% thus to enumerate all the frequent sequences that satisfy the user
% constraints. Note that the frequency itself can be seen as a
% constraint over the coverage of the sequences:
% $$frequent(S, \mathcal T) \equiv |cover(S, \mathcal T)| \ge minsup$$

%\subsection{Input}
% \begin{itemize}
% \item The vocabulary V : $\{v_0,  \ldots v_{|V|}\}$, with  $v_0 = \epsilon$
% \item $\mathcal{T}$: a set of $n$ transactions $\{T_1, T_2, \ldots ,T_n\}$, where a transaction
% $T_i$ is a sequence of characters of $V$.
% \item $supMin$: an integer $(\leq n)$.
% \end{itemize}

% \subsection{Output}
% All the frequent pattern $M$ of length at most $K$ that is to say the sequences of at most $s$ characters of $V$,
% which are subsequences\footnote{$seq'$ is a subsequence of $seq$ if $seq'$ can be derived from $seq$ by deleting some elements without changing the order of the remaining elements.} of at least $supMin$ transactions of $\mathcal{T}$. 
% TODO parler des fermés

%Syl: perdu le saut à la ligne

\section{SAT encoding of the base model}
\label{sec:model}
%\todo{pharse intro}
%\todo{uniformiser pattern sequence}

In this section, we present the variables and constraints of our SAT model.
We then show how to slightly modify the SAT solver in order
to generate all the frequent sequences. 

\subsection{Variables}
%\subsection*{Variables:} %Report to Figure \ref{fig:var}

\tikzstyle{block} = [draw,minimum size=2em]
% diameter of semicircle used to indicate that two lines are not connected
\def\radius{.7mm} 
\tikzstyle{branch}=[fill,shape=circle,minimum size=3pt,inner sep=0pt]

In frequent sequence mining, patterns (i.e., frequent sequences) can
have varied lengths, whereas constraint-based solvers usually require
a fixed number of variables. Therefore, one must first compute an upper bound
for the size of the patterns. Obviously, a pattern cannot be longer than
the longest transaction, but we may compute a tighter bound: to be
frequent, a pattern has to be covered by at least $minsup$
transactions. Therefore, the maximal size $K$ of a pattern is the
 size of the $minsup$ longest transactions in $\T$.  We then
define every possible pattern $S$ as a sequence of exactly $K$
characters and introduce an extra character $\epsilon$ in the
vocabulary $V$ to encode shorter patterns. $\V = V \cup \{\epsilon\}$.
For example, if $K = 4$, the pattern $ab$ will be represented as
$S=ab\epsilon\epsilon$.

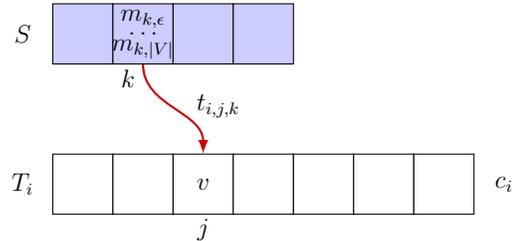
\begin{figure}[h]
\center{\large
        \begin{tikzpicture}[transform shape, scale=0.8]
        \tikzstyle{seq_item}=[draw, fill=blue!20, shape=rectangle, minimum size=1cm];
        \tikzstyle{seq_item_static}=[draw, fill=white, shape=rectangle, minimum size=1cm];
        \tikzstyle{match}=[out=270, in =90, thick, -latex, color=red!80!black];
       
        \node at (0, 0) {$S$}; 
        \setcounter{x}{1}
        \foreach \i in {1, 2, 3, 4}{
          \node[seq_item] at (\arabic{x}, 0) (p\arabic{x}) {};
          \stepcounter{x}
        }
        \node[yshift=-0.75cm, xshift=-0.25cm ] at (p2) {$k$};

         \node[yshift=0.25cm] at (p2) {$m_{k,\epsilon}$};
        \node at (p2) {$\ldots$};
        \node[yshift=-0.25cm] at (p2) {$m_{k,|V|}$};

  \node at (0, -2.5) {$T_i$}; 
        \setcounter{x}{1}
        \foreach \i in {B, A, C, B, C, A, C}{
          \node[seq_item_static] at (\arabic{x}, -2.5) (t1\arabic{x}) {};
          \stepcounter{x}
        }
        \draw (p2) edge[match] (t13);
        \node[yshift=-0.75cm ] at (t13) {$j$};
     
        \node at (8, -2.5) {$c_i$};

  %\node at (p2) {$m_{k,v}$};
  \node at (t13) {$v$};
  \node[yshift=1.3cm, xshift=0.25cm] at (t13) {$t_{i,j,k}$}; 
       
\end{tikzpicture}
}
\caption{\label{fig:vars}Variables of the model}
%\vspace{-1cm}
\end{figure}

We can now build our formulation around 3 sets of
variables: the $m$, $c$ and $t$ variables.
See Figure~\ref{fig:vars}.\\
%it would be nice to have a linespace here

\noindent\textit{$m$ variables:} For each position $k \leq K$ in the pattern $S$ and for each character $v$ of  the vocabulary $\V$,
we introduce a literal $m_{k,v}$, which is true if and only if the $k^{th}$ character 
of the pattern is equal to $v$: 
$$m_{k,v}=\top \Leftrightarrow S[k]=v$$
%($K \times |V| variables$).
\textit{$c$ variables:} For each transaction $T_i$, we add a literal $c_i$, which is true only if  the transaction $T_i$ is 
covered by the pattern:  
$$c_i = \top \Rightarrow  S \sqsubseteq T_i $$ 
\textit{$t$ variables:} For each transaction $T_i$, for each position $j$ in $T_i$ and for each position $k \leq j$ in $S$,
we add a literal $t_{i,j,k}$, which is true if  the $j^{th}$ character 
of the transaction $T_i$ is used as support for the $k^{th}$ character of $S$.
If $c_i =\top$, there exists, by definition  an embedding $e$ such that $S \sqsubseteq_e T_i$.
If $S \sqsubseteq_e T_i$ then $t_{i,j,k} = \top$ only if $e[k]=j$.
%$t_{i,j,k}$ requires $T_i[j] = S[k]$ to be true, but this is not the only requirement.    
$$t_{i,j,k} = \top ~\land~ S \sqsubseteq_e T_i   \Rightarrow  e[k]=j$$

Note that we use an implication rather than an equivalence %double implication %can we say a single implication?
because there may be multiple matches for the character $S[k]$ in the
transaction $T_i$. 
In Section \ref{sec:}, we explain how extra user constraints prevent us from selecting matches {\em a priori}.

\subsection{Constraints}
\paragraph{\textbf{Well-formed pattern}}
%Direct encoding rémi 
The two following  constraints encode the fact that each position $k$ in the pattern
can only be assigned exactly one character from the vocabulary. 
Constraint (\ref{eq:wellform1}) imposes
that {\em at least one} character is assigned per position  since constraint (\ref{eq:wellform2})  imposes that {\em at the most one} character is assigned per position:
{\small
\begin{equation}\label{eq:wellform1}
\forall k \in 1..K, (m_{k,\epsilon} \lor m_{k,1}\lor \ldots \lor m_{k, |V|}) 
\end{equation} 
}
%
%\begin{multline}\label{eq:wellform2}
%\forall k \in \{1,\ldots ,K\}, \forall v \in \{0,\ldots, (|V| - 1)\},\\ \forall v' \in \{(v+1), \ldots, |V|\},\ \ (\overline{m_{k,v}} \lor \overline{ m_{k,v'}})
%\end{multline} 
{\small
\begin{equation}\label{eq:wellform2}
%\forall k \in 1..K, \forall v \in 0..(|V| - 1),\\ \forall v' \in \{(v+1), \ldots, |V|\}, (\overline{m_{k,v}} \lor \overline{ m_{k,v'}})
\forall k \in 1..K, \forall v \in 1..V\mbox{-}1, \forall v' \in v\mbox{+}1..V,\ \ (\overline{m_{k,v}} \lor \overline{ m_{k,v'}})
\end{equation} 
}%
To avoid pattern symmetries, we discard every solution in
which at least one $\epsilon$ character does not appear at the end of the sequence. 
In other words, we forbid solutions in which $\epsilon$ is followed by any %syl
character $c \in V \setminus \{\epsilon\}$. % $m_{k,\epsilon} \implies m_{k+1,\epsilon}$:
%{\small
\begin{equation}\label{eq:symetries}
%\forall k \in \{1,\ldots ,(K-1)\},\\ (\overline{ m_{k,\epsilon} } \lor m_{k+1,\epsilon})
\forall k \in 1..K\mbox{-}1, (\overline{ m_{k,\epsilon} } \lor m_{k+1,\epsilon})
\end{equation} 
%}%
\paragraph{\textbf{Support compatibility}}
This constraint imposes that positions in the pattern can only be
mapped to the positions in the transaction if the characters are equal.
% For each transaction $T_i$, for each slot $j \leq |T_i|$, let $v$ be the (fixed) character $T_i[j]$,
% for each slot $k \geq j$
% We then may have $T_i[j] = M[k]$ ($t_{i,j,k} = \top$) only if $M[k] = v$ $(m_{k,v}=\top)$:
%
{\small
\begin{multline}\label{eq:compatibility}
%\forall i \in \{1, \ldots, |\T|\}, \forall j \in \{1,\ldots , |T_i|\}, \forall k \in \{1, \ldots, i\}\\  (\overline{ t_{i,j,k} } \lor m_{k,v}), \mbox{    where } v = T_i[j] 
\forall i \in 1..|\T|, \forall j \in 1..|T_i|, \forall k \in 1..i\\ (\overline{ t_{i,j,k} } \lor m_{k,v}), \mbox{    where } v = T_i[j] 
\end{multline} 
}

\paragraph{\textbf{Coverage constraint}}
%At least one valid support per slot in the pattern:
For each transaction $T_i$, 
$S \sqsubseteq_e T_i$ is  covered by the pattern ($c_i = \top$) only if, for each position $k$
of the pattern, either $S[k]=\epsilon$ ($m_{k,\epsilon} = \top$) or there exists a position $j$ such that
$e[k] = j$.
% \geq k$ in $T_i$ such that $T_i[j] = S[k]$: 
%
{\small
\begin{multline}\label{eq:coverage1}
\forall i \in 1..n, \forall k \in 1..K\mbox{-}1,  (\overline{ c_i} \lor m_{k,\epsilon}\lor t_{i,k,k}\lor \ldots  \lor t_{i,|T_i|,k})
%\forall i \in \{1, \ldots, n\}, \forall k \in \{1, \ldots, K\},\\  (\overline{ c_i} \lor m_{k,\epsilon}\lor t_{i,k,k}\lor t_{i,k+1,k} \lor \ldots  \lor t_{i,|T_i|,k})
\end{multline} 
}%
To avoid symmetries in embeddings, each character in the pattern can match only once in the transaction. 
%For each transaction $i$, for each position $k$ of the pattern, it exists at most one position $j$, which is a valid support for $M[k]$.
%
\begin{multline}\label{eq:coverage2}
%\forall i \in \{1 \ldots n\},
%\forall k \in \{1 \ldots (K - 1)\},\\
%\forall j \in \{1, \ldots, (|T_i| - 1)\},
%\forall j' \in \{(j+1)\ldots |T_i|\},\\
%\quad \quad (\overline{t_{i,j,k}} \lor \overline{t_{i,j',k}})
\forall i \in 1..n, \forall k \in 1..K\mbox{-}1,\\
\forall j \in 1..|T_i|\mbox{-}1,
\forall j' \in j\mbox{+}1..|T_i|,
(\overline{t_{i,j,k}} \lor \overline{t_{i,j',k}})
\end{multline} 

\paragraph{\textbf{Order preservation constraint}}
For each transaction $T_i$, for each position $k$ of the pattern, position $j$ is a valid support for $S[k]$ 
($t_{i,j,k} =\top$) only if it exists $k-1\leq j' <j$ s.t. position $j'$ is a valid support for $S[k-1]$ ($t_{i,j',k-1} =\top$): 
%
%$t_{i,j,k} \implies \exists j' < j ~s.t.~ t_{i,j',k-1}$ encoded in 

\begin{multline}\label{eq:preserveorder}
%\forall i \in \{1, \ldots, n\},\forall j \in \{1,\ldots, |T_i|\}, \forall k \in \{1, \ldots, j\},\\
\forall i \in 1..n,\forall j \in 1..|T_i|, \forall k \in 1..j,\\
(\overline{t_{i,j,k}} \lor t_{i,k-1,k-1} \lor t_{i,k,k-1} \lor \ldots \lor t_{i,j-1,k-1} )
\end{multline} 

\paragraph{\textbf{Cardinality constraint}} The last constraint of the sequence mining problem is 
the minimum frequency threshold $minsup$. To ensure that pattern $S$ is frequent enough, we need
to encode the requirement that at least $minsup$ transactions $T_i \in \T$ are covered ($c_i = \top$).

\begin{equation}\label{eq:frequent}
atLeast(minsup, [c_1,\ldots,c_{|\T|}])
\end{equation} 
%
%%Since this constraint appears in numerous industrial applications,
%%several formulations have been proposed already. We briefly summarize
%%this work. Cardinality constraints can be directly expressed in
%%pseudo-boolean solvers, such as \cite{sat4j} or they can be decomposed
%%into a pure SAT formulation using extra literals. For example: cardinality network
%%decomposition \cite{BailleuxCP03} allows decomposition using
%%$\bigO(|\T|.log^2(|\T|))$ extra literals. This number of extra
%%literals has been later improved to $\bigO(|\T|.log^2(minsup))$ in
%%\cite{SinzCP05}, and show as optimal.
%%%%
%%More recently, Abio and Stuckey \cite{StuckeyCP12} proposed a
%%technique to improve the efficiency of the cardinality constraint, by
%%taking into account two tricky internal mechanisms of a CDCL SAT
%%solver, namely the explanation of conflicts and the activity of
%%variables after the restarts.
%
This constraint appears in numerous industrial applications, 
and several decompositions in CNF have been proposed in the literature. 
%edinto a pure SAT
%several formulations have been proposed already. We briefly summarize
%this work. Cardinality constraints can be directly expressed in
%pseudo-boolean solvers, such as \cite{sat4j} or they can be decomposed
%into a pure SAT formulation using extra literals. For example: 
Bailleux and Boufkhad \cite{BailleuxCP03} proposed a network-based 
decomposition using $\bigO(|\T|.log^2(|\T|))$ extra literals. 
This number of extra literals was later reduced to $\bigO(|\T|.log^2(minsup))$ in
\cite{SinzCP05}, and proved optimal. %Syl: as optimal?
More recently, Abio and Stuckey \cite{StuckeyCP12} have proposed a
technique to improve the efficiency of the cardinality constraint, by
taking into account two tricky internal mechanisms of a CDCL SAT
solver. Specifically, the authors suggested using the explanation of conflicts
and the activity of variables after the restarts.
%
%Remi : re-écrire le paragraphe suivant  
However, in some variants of our problems (presented in Section
\ref{sec:solv-deriv-sequ}), the cardinality between $minsup$ and $|\T|$ must 
be preserved. And because of the logarithmic formulation,
there is no simple way to express sub-cardinalities in these
approaches. We therefore used the formulation initially proposed by Coquery et
al. in \cite{lakdar12}. Although this decomposition involves more variables
$\bigO(|\T|^2)$, it allows the simple expression of cardinality. 
Indeed, when building the decomposition for $|\T|$, we obtain a
literal $card_t$ per value $t \in minsup..|\T|$, stating that if
$card_t$ is true, there exist at least $t$ literals among
$c_1,\ldots,c_{|\T|}$ which are true. %syl: revoir avec Rémi.

%%remi : on vire pas de place
%\subsection{Spacial complexity}
%Constraint \ref{eq:wellform1} $\bigO(K)$ clauses of size $\bigO(V)$.
%\ref{eq:wellform2} $\bigO(|V|^2)$ binary clauses.
%\ref{eq:symetries} $\bigO(K)$ binary clauses.
%\ref{eq:compatibility} $\bigO(n\times l \times K)$ binary clauses. 
%\ref{eq:coverage2} $\bigO(n \times l^2)$ binary clauses
%\ref{eq:preserveorder} $\bigO()$
%\ref{eq:frequent} $\bigO(|\T|^2)$ variables and binary clauses. 

\subsection{Solving}
\label{sec:solving}
%\todo{THEOREME:  models(F) = patterns frequent}

We solve our SAT formula using the Glucose CDCL solver
\cite{Glucose09}. Truc et Machin have shown in \cite{GiunchigliaMT02} %remi
that branching over decision variables confuses the clause-learning %branching on over ? 
system and ultimately reduces the solving performances. Therefore, we
do not provide any hint for the search strategy and use the solver
as a black box.

Each model of our SAT formula (i.e., each solution) is a frequent
sequence of the input dataset. As a result, outputting all the frequent
sequences can be done by enumerating all the models that satisfy the
formula.

%   . The main explanation is that one of
% the interest of SAT solvers consists in letting it do the job of
% finding difficult parts of the problems. Such difficult parts may
% involve auxiliary variables and preventing the solver to branch over
% these variables then often results in decreasing the performance of
% the search.
%
\begin{scriptsize}
\begin{algorithm}[h]
\SetKwInOut{Input}{input}
\SetKwInOut{Output}{output}
%\Input{SAT formula F}
\nl \While{$F$ is solvable} { \label{nl:while}

\nl $model \leftarrow nextModel(F)$ \label{nl:next}
 
\nl 	
 $F \leftarrow F \land (\bigvee {\overline{m_{k,v}} ~s.t.~ model[m_{k,v}] = \top \land v\neq \epsilon})$ \label{nl:encode}
%	  \overline{m_{1,v}}  \lor \ldots \lor \overline{m_{K,v}} ~, ~with ~model[m_{k,v}] = \top ) $
}
\caption{\textsc{Compute all frequent patterns}}
\label{algo:solveAll}
\end{algorithm}
\end{scriptsize}

However, more specifically, only the literals $m_{k,v}$ of the model 
correspond to a frequent sequence; indeed, literals $c_{i}$ 
represent the set of covered transactions, and literals $t_{i,j,k}$
an embedding of this sequence in each covered transaction.
Since each covered transaction may have several embeddings,
the number of solutions for our SAT formula may be exponentially 
larger than the number of frequent sequences. 

To eliminate this potential issue, we propose Algorithm~\ref{algo:solveAll} as a simple way to enumerate all the models: we first solve the formula and find a model (line 2);   
we then restrict the model to literals encoding the frequent sequence, and subsequently post the negation of the restricted model as a new clause (line 3). The process is repeated
 until no additional model can be found (line 1).

%
%Each model of our SAT formula (i.e. each solution) is a frequent
%sequence in the input dataset, therefore outputting all the frequent
%sequences can be done by enumerating all the models that satisfy the
%formula. A simple way to enumerate all the models is to 1) solve the
%formula and find a model, 2) post the negation of the model as a new
%clause and 3) repeat this process until no more model can be found.
%However, this is a complete but incorrect way to enumerate the set of
%frequent sequences because several models can code for the same
%frequent sequence (two models can code for the same pattern with two
%different embeddings). In order to avoid duplicates sequences, we only
%post the negation of the pattern variables as a new clause
%(i.e. $m_{k,v}$ variables). This not only prevent from enumerating the
%same model but also all the models that code for the same sequence.
%This technique is described in Algorithm~\ref{algo:solveAll}.
%
%

% In order to enumerate all sequences and not just one, we post the
% negated sequence every time a model is found 
% In a very classical way when looking for all solutions with a CDCL solver, 
% each time a model is found, its negation as a new clause of the formula 
% to avoid the solver to reach it another time.
%   Glucose consists in the way
% we encode the no-goods representing each solution. First of all, we
% are only interested in the literals encoding the pattern, namely the
% $m_{k,v}$. Indeed, a given pattern (assignment of the $m_{k,v}$) may
% lead to several embedding for each transaction it covers.  

\section{Extending the model with user constraints}
\label{sec:solv-deriv-sequ}
Enumerating all frequent sequences may yield an overwhelming number %yield = engendrer %leads to : conduire à
of sequences. (See second column in Table~\ref{tab:spec}.)
Some of them may be deemed redundant and/or irrelevant by the
user.  In
this section, we formulate various user constraints that can be added
to the base model in order to extract fewer and more specific sequences. The following
is not an exhaustive list of useful user constraints, but rather a series of
examples that demonstrate the flexibility of the SAT formulation.
% to address various sequence mining tasks derived from the standard
% sequence mining task defined in Section~\ref{sec:freq-sequ-mining} to
% enumerate fewer and more interesting sequences.

\subsection{Constraints on the way a pattern is embedded}
\label{sec:userembed}

\paragraph{\textbf{Sequence mining with maximum gap constraint}} With
the maximum gap constraint, transactions are covered only if the
characters of the sequence occur in the transaction with a gap of at
most $\gamma$ characters. We define the $\maxgap$ predicate over an
embedding $e = (k_1,\ldots,k_n)$ as follows:

\begin{equation}
  \label{eq:gaps}
  \maxgap_\gamma(e) \equiv \forall i \in 1\ldots n\mbox{-}1, (k_{i+1} - k_i) \le \gamma  
\end{equation}

  To expand the standard sequence mining problem with this constraint,
  we only have to append the following constraint to the formula. 

%To enforce this constraint in our SAT formula, we simply have to slightly  modify the Preserve-Order constraint (\ref{eq:preserveorder})

%$t_{i,j,k} =1 \implies \exists  max(k-1,j-gap_k) < j' < j ~s.t.~ t_{i,j',k-1} =1$   encoded in
  This constraint imposes that the distance between indices $j$ and $j'$
  of $t_{i,j,k}$ and $t_{i,j',k}$ is smaller than $\gamma$:
  \begin{equation}
    \label{eq:gapsformula}
    (\overline{t_{i,j,k}} \lor t_{i,max(k-1,j-\gamma),k-1} \lor \ldots \lor t_{i,j-1,k-1})    
  \end{equation}

%\paragraph{Remark on the gap constraint} Due to its usefulness, the
%  gap constraint has been implemented in several specialized sequence
%  mining algorithm but with a different sementic. For example the
%  LCMseq algorithm~\cite{OhtaniKUA09} can mine frequent sequence with
%  gap constraints, but the gap constraint is only checked
%  on {\em the first embedding} of the sequence in each transaction. \noindent We illustrate on the following dataset, with $minsup = 2$ and $max\mbox{-}gap = 2$: \\
%  $\begin{array}{l l}
%    T_1: & A C C B A B \\
%    T_2: & A B
%\end{array}$
%
%\noindent In this dataset, LCMseq does not count $A B$ as a frequent
%sequence because the first embedding of $A B$ in $T_1$ ($A_1 B_6$)
%does {\em not} satisfy the gap constraint. Counting all the
%transaction that contains {\em at least one} embedding that satisfies
%the gap constraint (as formulated in Equation~\ref{eq:gaps}) is more
%natural, but it also requires a more complex algorithm to {\em search}
%for an embedding satisfying the gap. Constraint based solvers are a
%very natural way to solve decisions problems like this. Indeed, in contrast
%with specialized algorithm SAT solvers natively support this
%constraint.

%
  \paragraph{Remark on the gap constraint} Due to its usefulness, the
  gap constraint has already been implemented in 
  LCMseq algorithms~\cite{OhtaniKUA09}.
  Unfortunately, though correct, this implementation is not complete.  
   The gap constraint seems to be checked only on {\em the first embedding} of 
   the sequence in each transaction. \noindent We illustrate this shortcoming with the following dataset, using $minsup = 2$ and $max\mbox{-}gap = 2$: \\
  $\begin{array}{l l}
    T_1: & A C C B A B \\
    T_2: & A B
\end{array}$

\noindent In this dataset, LCMseq is unable to find $A B$ as a frequent
sequence because the first embedding of $A B$ in $T_1$ ($A_1 B_6$)
does {\em not} satisfy the gap constraint. \\

Indeed, counting all the
transactions that contain {\em at least one} embedding satisfying
the gap constraint (as formulated in Equation~\ref{eq:gaps}) requires a more 
complex algorithm.% to {\em search} for embeddings satisfying the gap. 
This short example points out the difficulty to integrate extra constraints in a 
%DataMining specialized 
traditionnal mining algorithms. %Syl: DataMining in one word + capitals? to search for such embeddings?

    In contrast, our approach, based on an SAT solver, tackles this issue in a sound and complete way.
% to solve decisions problems like this. 
%Indeed, in contrast
%with specialized algorithm SAT solvers natively support this constraint.
%Syl: perdu le saut à la ligne

\paragraph{\textbf{Sequence mining with dependent gap constraint}} With the generalized
  gap constraint, transactions are covered only if the two consecutive sequence characters occur in the transaction separated by a gap depending on the position in the sequence %Syl: whose position?
  and/or the character preceding the gap. More formally, given a function 
  $gap : (K \times V) \rightarrow [1..l]$ , we define the $\gengap$ predicate over an embedding $e =
  (k_1,\ldots,k_n)$ as follows:
  $$\gengap_{gap}(e) \equiv \forall i \in 1\ldots n\mbox{-}1, (k_{i+1} - k_i) \le gap(i,S[i])$$
This constraint may be enforced by %the implication 
%$t_{i,j,k} =1 \land m_{k,v} \implies \exists  max(k-1,j-gap(k,v)) < j' < j ~s.t.~ t_{i,j',k-1} =1$
%encoded in 
the following clause:
$$(\overline{t_{i,j,k}} \lor (\overline{m_{k,v}} \lor t_{i,max(k-1,j-gap(k,v)),k-1} \lor \ldots \lor t_{i,j-1,k-1})$$

\paragraph{\textbf{Sequence mining with maximum span constraint}} With the maximum
  span constraint, transactions are covered only if the first and last characters of the sequence 
  occur in the transaction separated by at most $\gamma$ characters. More formally, we define the 
  $\maxspan$ predicate over an embedding $e = (k_1,\ldots,k_n)$ as follows:
  $$\maxspan_\gamma(e) \equiv \forall i \in 1\ldots n\mbox{-}1, (k_{n} - k_1) \le \gamma$$
To encode this constraint, we define two additional literals per character in each transaction:
$f_{i,j}$ points to the position of the support for the first character of $S$, and $u_{i,j}$
is true for each position $j$ in the transaction $T_i$ between the support for the first and the 
last character of $S$. 
$f_{i,j}$ and $u_{i,j}$ are subjected to the following constraints: %Syl: are ?
{\small
$$\left\{
\begin{array}{lr} 
    \forall i\in1..|\T|, \forall j\in1..|T_i|, (\overline{t_{i,j,0} } \lor  f_{i,j}) \land (t_{i,j,0} \lor  \overline{f_{i,j}}) &(i) \\   
    \forall i\in1..|\T|, \forall j\in1..|T_i|, \forall k\in 1..j  (\overline{t_{i,j,k} } \lor  f_{i,j}) &(ii)\\
    \forall i\in1..|\T|, \forall j\in1..|T_i|, (\overline{u_{i,j}} \lor f_{i,j} \lor u_{i,j-1} &(iii)\\
    \forall i\in1..|\T|, \forall j\in1..|T_i|-span, (\overline{c_i} \lor \overline{u_{i,j}}  \lor \overline{u_{i,j+span}}) &(iv)
\end{array}\right.
$$
}
%
%\todo{Remi, reformules le paragraphe suivant? Je comprend pas ce que tu veux dire.}
%Fix July 20 th  
%
%remi revoir ce paragraphe !
Clauses (i) ensure that if $Ti[j]$ is used as the support for the first position of sequence ($t_{i,j,0} = \top$), then is marked as the first support $f_{i,j} = \top$; clauses (ii) mark  any position $k$ in the sequence ($\forall t_{i,j,k} = \top$, we have $u_{i,j}  = \top$);
clauses (iii) enforces positions $i'$, which are not used as a support themselves but located between the first support $f_{i,j}$ and  support and another support, to also satisfy $u_{i',j}$. Finally, (iv) enforces a gap of at most $span$ characters between the support of the first and the last character of $S$.

%Syl: review above paragraph with Remi
% \paragraph{\textbf{sequence mining with average gap constraint:}} we also propose an
%   average gap constraint in which transactions are covered in the
%   frequency count only if the characters of the sequence occur in the
%   transaction with an average gap of at most $\alpha$ characters.
% More formally, we define the $\avggap$ predicate over an embedding $e =  (k_1,\ldots,k_n)$ as follows:
%   $$\avggap_\alpha(e) \equiv \frac{(k_{n} - k_1)}{n-1} \le \alpha$$
% %
% This $\avggap$ constraint may be combined with the $\maxgap$ constraint, for instance  allowing 
% a max gap of 3 but an average gap of 2.
% %
% The SAT formulation of this constraint is very similar to $\maxspan$, where the (iv) clause is replaced 
% by $K$ clauses: 
% $$\forall i\in1..|\T|, \forall k\in 1..K \forall j\in1..|T_i|-\gamma\times k, (\overline{c_i} \lor \overline{u_{i,j}} \lor \overline{m_{k,\epsilon}} \lor \overline{u_{i,j+\gamma\times k}})$$
% %
% %    The goal is then to extract all the sequences $S$ such that 
% %  $$|\{T : \exists e\mbox{ s.t. }S \sqsubseteq_e T
%  % \land \avggap_\alpha(e)\}| \ge minsup$$

% %For instance, we may want to constraint the gap between two supports in a transaction:
% %For instance, let $ab$ be a pattern and the maximal gap $g= 2$ the max allowed gap, $acb$ is covered by the pattern since $acccb$ is not.
% %Such a constraint may be encoded by refining the preserve order constraint  in the following clauses:

\subsection{Constraints over the pattern itself}
\label{sec:userpattern}

In many application fields, interesting sequences can be distinguished
from irrelevant ones by specifying syntactic constraints on the
pattern itself. For example, the authors of \cite{albert2003framework}
designed an algorithm to mine frequent sequences that satisfy regular
expressions specified by the user. The resulting algorithm is applied
to biological datasets.

We now explain how to address this type of constraint in our SAT
formulation. In \cite{pesant04}, Pesant proposed the {\em regular}
global constraint for CP systems. Indeed, this constraint is satisfied if a
sequence of values belongs to a given regular language. %Syl: what's a regular language?  
More recently, in \cite{quimperw06,sellmann06}, the authors extended this work to the
context-free languages with the introduction of {\em grammar} global
constraints.
Furthermore, Quimper and Walsh
\cite{QuimperGrammar08} %Syl: the reference doesn't show in the text.
and by Katsirelos et al.~\cite
{KatsirelosGrammar09} have shown how to decompose regular and grammar 
%(an extension of context-free languages) 
constraints into SAT formulae. Based on their work, 
we can easily incorportate expression and grammar constraints into our SAT formulation.

\section{Maximal and closed pattern mining}
\label{sec:maxim-clos-patt}

% \paragraph{solving derived sequence mining tasks:}
% \label{sec:solv-deriv-sequ}

In our SAT formulation and in other CP/SAT formulations of 
pattern mining problems (e.g., \cite{DeReadt08}), each solution represents a pattern.
Usually, in classical SAT or CP applications, satisfiability (i.e., finding one solution only) is the main concern.
Conversely, pattern mining users are not interested in finding a single solution, but multiples ones.
 In Section~\ref{sec:solving}, we showed how to enumerate
all the solutions with minor modifications in the SAT solver. 

%Moreover, pattern mining users have more specific needs. 
However, the set of all patterns is usually highly redundant. To overcome this issue, %Syl: review last 2 sentences with Remi.
condensed representations were introduced. For example, closed pattern
mining~\cite{pasquier1999closed} aims at extracting a subset of patterns from which it is possible to derive all other
patterns. Machin et Truc \cite{bonchi2004closed} later showed  that expressing
closedness as a constraint can lead to ambiguous problem definition.
In this section, we address closed and maximal pattern
mining~\cite{maximal98} by using interactive SAT solving.

% In practice, enumerating all the frequent sequences can lead to an
% overwhelming number of sequence that are sometimes redundant and/or
% not relevant to the user. A very common way to address this problem in
% pattern mining is to derive a new problem definition to enumerate
% fewer and more interesting patterns. However, there are many possible
% problem definition and the choice of the adequate problem definition
% usually depends on the user needs and the problem at hand.

% A very common way to address this problem in pattern mining is
% to derive a new problem definition to enumerate fewer and more
% interesting patterns. % However, there are many possible problem
% % definition and the choice of the adequate problem definition usually
% % depends on the user needs and the problem at hand.

\paragraph{\textbf{Closed sequence mining}} This was first introduced by
Pasquier et al. for itemset mining
(\cite{pasquier1999closed}). 

Given the frequent sequence $S = v_1v_2\ldots v_n$, we say that $S$ is
closed if there is no character $v'$ such that $S' = v_1v_2\ldots v_nv'$
has the same coverage.
  $$closed(S,\T) \equiv$$$$ \nexists v' \in V\mbox{ s.t. } cover(S,\T) = cover(Sv',\T)$$

In our base formulation, constraints \ref{eq:compatibility} and \ref{eq:preserveorder}
prevent the validity of supports $t_{i,j,k}$ due to the compatibility and the %Syl: Review with Remi
order of supports. Then, constraint \ref{eq:coverage1} prevents the $c_i$ from being satisfied, 
due to the validity of supports $t_{i,j,k}$ ($c_i$ only appears as a negative literal).
The only constraint that mandates that $c_i$ be satisfied is the frequency constraint %remi: pas : mandates c_i to be satisfied
\ref{eq:frequent}. As a result, given a pattern $S$, if $|cover(S,\T)| > minsup$,
there may exist transactions $T_i$ such that $S \sqsubseteq T_i$ but $c_i=\bot$.
This is captured by the definition of $c_i$ given in Section \ref{sec:model}.
Unfortunately, the definition of $closed(S)$ requires the maximization of
$|cover(S,\T)|$, which would only be possible if we had $c_i =\top
\Leftrightarrow S \sqsubseteq T_i$ instead of $c_i =\top \Rightarrow S
\sqsubseteq T_i$.
However, coverage, order preservation and other user-defined constraints
(\textit{see} Sections \ref{sec:userpattern} and \ref{sec:userembed})
only impose restrictions on supports. Having $c_i =\top
\Leftrightarrow S \sqsubseteq T_i$ would require that we combine the
negation of each of these constraints, then build a DNF;
this would result in an exponential number of clauses.

To avoid this constraint negation requirement, we now propose to use the incremental SAT solving paradigm %under certain assumptions
\cite{SATAssumption12}.
Algorithm \ref{algo:solveAllClosed} performs several calls to the SAT solver of formulae %Syl: perform a call?
sharing the same set of clauses, but with a different set of assumptions.

The main advantage of SAT solving under assumptions is that the
clauses learned during a solving procedure under a given set of %Syl: learned = terme consacré?
assumptions are kept for later solving. The learned clauses
are removed only if they involve a literal that occurs in an assumption
that is also removed.

Since nested  loops (lines \ref{ln:Ioop1} and \ref{ln:Ioop2}) ensure that for any pair 
of patterns $S_1$ and $S_2$ found, with  $S_1$ found before $S_2$, we have 
either $covers(S_1,\T) \subset covers(S_2,\T)$ or $covers(S_1,\T) = covers(S_2,\T)$ and $|S_1| \geq |S_2|$.
For each pattern $S$ found, line \ref{ln:Ioop3} adds no-goods corresponding to each %syl: no-goods?
subsequence $S' \sqsubseteq S$ to avoid generating $S'$ in a future resolution.

\begin{algorithm}[h]
{\footnotesize
%\dontprintsemicolon
\SetKwRepeat{Repeat}{do}{while}
\SetKwInOut{Input}{input}
\SetKwInOut{Output}{output}

$Stack: A \leftarrow \{card_{minsup+1},\ldots , card_{|T|}\}$

\lnl{ln:Ioop1} \Repeat{$A\neq \emptyset$} { 
  $Stack: A' \leftarrow \{\overline{m_{1,\epsilon}},\ldots , \overline{m_{K,\epsilon}} \}$\;
\lnl{ln:Ioop2}  \Repeat{$A'\neq \emptyset$} {
\lnl{ln:solveloop} 	 \While{$F$ is solvable under assumption($A \land A'$)} {
		$model \leftarrow nextModel(F)$\;
\lnl{ln:Ioop3}\ForEach{$pattern \in \mathcal{P}(m_{k,v} ~s.t.~ model[m_{k,v}] = \top ~and~v\neq \epsilon\})$} {	
			 $F \leftarrow F \land (\bigvee_{m_{k,v} \in pattern}~ \overline{m_{k,v}}  \lor \overline{m_{|pattern|,\epsilon}}) $
	 	}
	}
 	$A'.pop()$	
  }
  $A.pop()$
}
}
\caption{\textsc{Compute all closed patterns}}
\label{algo:solveAllClosed}
\end{algorithm}

%\todo{THEOREM: correctness and completeness}

\todo{PHRASE DE LIAISON}

\paragraph{\textbf{Maximal sequence mining}} Mining maximal sequences
aims to extract only the largest sequences that remain %Syl: if they're frequent, they don't "remain". rather, they "occur".
frequent. This further reduces the number of generated sequences, but
the set of all frequent sequences cannot be derived from the set of
maximal sequences.

    $$maximal(S,\T) \equiv$$$$ \nexists v' \in V\mbox{ s.t. } |cover(S.v',\T)| \ge minsup $$

    We address the problem of finding maximal sequences with a minor
    modification in Algorithm~\ref{algo:solveAllClosed} by only
    removing the external loop; indeed, we are not concerned with maximizing the
    number of supported transactions as we were for closed patterns.

\section{Experiments}
\label{sec:experiments}

After presenting our experimental setting in
Section~\ref{sec:experimental-setting}, we now compare our
approach to specialized algorithms designed to mine closed frequent
sequences. Subsequently, we demonstrate, that unlike specialized algorithms,
our SAT-based approach is able to take advantage of extra constraints to further
prune the search space.
%What is the impact on efficiency, of the optimizations that we have proposed in Section~\ref{sec:exploting}?
 
%The experiments that we conducted aim at answering the following two questions:
%\begin{description}
%\item[Question~1]  What is the impact on efficiency, of the optimizations that we have proposed in Section~\ref{sec:exploting}?
%\item[Question~2]  How does the solver behave when incorporating extra user constraints? Can the solver take advantage of these
%constraints to further prune the search space?
%\end{description}
%
%We present our experimental setting in
%Section~\ref{sec:experimental-setting}, and then answer these two
%questions in Section~\ref{sec:impact-optimizations}, and
%\ref{sec:adding-extra-constr} respectively.
%
\subsection{Experimental setting}
\label{sec:experimental-setting}

We run our experiments on two real datasets called {\em Gazelle} and
{\em JMLR}. {\em Gazelle} is a dataset of web clicks data, in which
each transaction represents the sequence of page views by a single
user. {\em JMLR} consists in a set of paper abstracts from the {\em Journal
  of Machine Learning Research} in which each transaction is the abstract
of a publication. Both datasets were previously used to evaluate
sequence mining algorithms, for example in
\cite{wang2004bide} or in
\cite{DBLP:conf/kdd/TattiV12}. Since our approach is currently unable
to tackle the entire datasets, we derive smaller datasets by taking
the first $n$ transactions. For example, the {\em JMLR-500} dataset
consists of the 500 first transactions of {\em JMLR}. This approach is
preferable to synthetic datasets, which usually exhibit very different
properties than real datasets. As shown in
Table~\ref{tab:datasets}, the properties of {\em JMLR-500} and {\em
  Gazelle-500} are similar to the ones of {\em JMLR} and {\em Gazelle}
respectively.

\begin{table}
  \centering
  \begin{tabular}{|l|c|c|c|c|}
    \hline
    Dataset & \# transactions & voc size & max  length & avg  length \\\hline
    {\em JMLR} & 788                &  3847                      & 231 & 96.997                      \\\hline
    {\em JMLR-500} & 500                & 3208                       & 231 & 95.754                      \\\hline
    {\em Gazelle} &         29369        &   1424                     &    651           & 2.981        \\\hline
    {\em Gazelle-500} &         500        &  1094                      &  651 & 15.052                     \\\hline
  \end{tabular}
  \caption{Charachteristics of {\em Gazelle} and {\em JMLR}. }
  \label{tab:datasets}
 \vspace{-1cm}
\end{table}

When the frequency threshold is lower, the number of sequences to
enumerate is larger. To solve our SAT formulation, we use the
Glucose\cite{Glucose09} SAT solver modified as explained in
Algorithm~\ref{algo:solveAllClosed}. We only mine for closed %Syl: we only mine
sequences, since non-closed sequences can be derived from closed ones.
%
 %  and
We run all our experiments on a Linux host with an Intel Core i7-2600 CPU
at 3.40GHz and 16~GiB of memory.

\subsection{Enumerating all closed frequent sequences}

Specialized algorithms such as 
LCMseq~\cite{OhtaniKUA09} or Bide~\cite{wang2004bide} have been
designed to efficiently enumerate all closed frequent sequences.
Run times for these algorithms are presented in
Table~\ref{tab:spec}.

Our SAT-based approach does not purport to be faster than algorithms designed and optimized specifically for this basic task. Even then,
it is worth noting that the number of closed frequent sequences grows
quickly and makes the manual interpretation of the results almost
impossible. (See Table~\ref{tab:spec}, second column). In
this case, reducing the number of results by adding constraints is the
only option offered to the user.

In some cases, post-processing the set of closed frequent sequences may
be an alternative to using our constraint-based sequence mining
framework. However, this option is not always available, and, when available, not always efficient.

For example, post-processing the pattern set in order to eliminate all
the sequences that do not satisfy a $\maxgap$ constraint is
non-trivial: one needs to compute all the embeddings in the input %Syl: non-trivial
dataset, and then check whether there exists one embedding satisfying
the $\maxgap$ constraint. Since the number of embeddings for a pattern
is combinatorial with the number of valid matches for each character, %syl: combinatorial with
there is no guarantee that this can be done in a reasonable amount of time. 
%remi: pourquoi c'est dans les expés !

Furthermore, the set of closed frequent sequences is not guaranteed to %Syl: which?
be a superset of  closed frequent sequences satisfying a
given constraint. As pointed by \cite{bonchi2004closed}, this is
true only for anti-monotonic constraints. In short, a frequent
pattern that is not closed may become closed if a constraint discards
its closed superset from the solution set. In this case,
post-processing is simply impossible and existing sequence algorithms
cannot be used.

%%\begin{table}
%%  \centering
%%  \begin{tabular}{|l|c|c|c|c|}
%%    \hline
%%    Dataset@support & \# clo. patterns & CloSpan & LCMseq & Bide \\\hline
%%%%    {\em JMLR} & 788                &  3847                      & 231 & 96.997                      \\\hline
%%%%    {\em JMLR-500} & 500                & 3208                       & 231 & 95.754                      \\\hline
%%%%    {\em Gazelle} &         29369        &   1424                     &    651           & 2.981        \\\hline
%%    {\em Gazelle-500@10\%}  &     1       & 0.03s                               &  0.01s     & 0.05s   \\\hline
%%    {\em Gazelle-500@5\%}   &     14      & 0.05s                               & 0.01s      & 0.03s \\\hline
%%    {\em Gazelle-500@1\%}   &     4405    & 0.11s                               &  0.20s     & 0.15s   \\\hline
%%    {\em Gazelle-500@0.5\%} &     373568  & $>$ 1000s                           &  $>$ 1000s & 26.54s   \\\hline\hline
%%    {\em JMLR-500@10\%}     &  636       & 0.09s  & 0.04 &  0.08s                                 \\\hline
%%    {\em JMLR-500@5\%}      &  4751     & 0.05s  & 0.04 &  0.17s                                       \\\hline
%%    {\em JMLR-500@1\%}      &  623011   & 9.00s  & 0.09 &  7.39s                                       \\\hline
%%    {\em JMLR-500@0.5\%}    &  2782799  & 64.33s & 0.61 &  51.76s                                  \\\hline
%%  \end{tabular}
%%  \caption{Number of closed patterns and runtime on existing algorithm for {\em Gazelle} and {\em JMLR}. }
%%  \label{tab:spec}
%%%  \vspace{-1cm}
%%\end{table}

\begin{table}
  \centering
  \begin{tabular}{|l|c|c|c|c|}
    \hline
    Dataset@support & \# clo. patterns  & LCMseq & Bide \\\hline
%%    {\em JMLR} & 788                &  3847                      & 231 & 96.997                      \\\hline
%%    {\em JMLR-500} & 500                & 3208                       & 231 & 95.754                      \\\hline
%%    {\em Gazelle} &         29369        &   1424                     &    651           & 2.981        \\\hline
    {\em Gazelle-500@10\%}  &     1                                      &  0.01s     & 0.05s   \\\hline
    {\em Gazelle-500@5\%}   &     14                                     & 0.01s      & 0.03s \\\hline
    {\em Gazelle-500@1\%}   &     4405                                   &  0.20s     & 0.15s   \\\hline
    {\em Gazelle-500@0.5\%} &     373568                             &  $>$ 1000s & 26.54s   \\\hline\hline
    {\em JMLR-500@10\%}     &  636         & 0.04 &  0.08s                                 \\\hline
    {\em JMLR-500@5\%}      &  4751       & 0.04 &  0.17s                                       \\\hline
    {\em JMLR-500@1\%}      &  623011    & 0.09 &  7.39s                                       \\\hline
    {\em JMLR-500@0.5\%}    &  2782799  & 0.61 &  51.76s                                  \\\hline
  \end{tabular}
  \caption{Number of closed patterns and runtime on existing algorithm for {\em Gazelle} and {\em JMLR}. }
  \label{tab:spec}
 \vspace{-1cm}
\end{table}

\subsection{Solving sequence mining with extra user constraints}
\label{sec:adding-extra-constr}

% The most important limitation of existing specialized algorithms is
% that they cannot be extended with arbitrary user constraints. As a
% consequence, a pattern mining practitioner that wants to mine sequence
% with a (possibly unknown) constraint has no other option than running a
% general sequence mining algorithm and then post process the generated
% patterns to extract the patterns that satisfy the constraint. This has
% two important drawbacks: first the sequence mining algorithm has to
% explore a search space that is uselessly large since it contains all
% the patterns that do or do not satisfy the extra user
% constraint. Second the post-processing can take some time. If the
% constraint is very restrictive the number of intermediate patterns can
% be overwhelming even through there are a small number of patterns
% satisfying all the constraints.

Next, we demonstrate that, unlike specialized
algorithms, the SAT solver is able take advantage of extra user
constraints to reduce mining time. To do so, we compare run
times based on various constraints:
\begin{itemize}
\item $\maxgap$ and $\maxspan$ constraints, which regulate the embedding of sequences in the transactions;
\item $regular$  constraints: for example, in the {\em JMLR} dataset, we extract the sequences matching the regular expression: 
 "$\star machine \star learning \star$"\footnote{
Note this very simple regular constraint only make sense when they are
combined with other constraints such as gap constraints. Otherwise a
simple pre-processing removing all the transactions not containing the
characters would be sufficient.}; 
\item a combination of the above constraints.
\end{itemize}

%
%\todo{Ajouter une r-expression avec disjonction? }

\begin{figure}[h]
 \centering
     \includegraphics[width=0.4\textwidth]{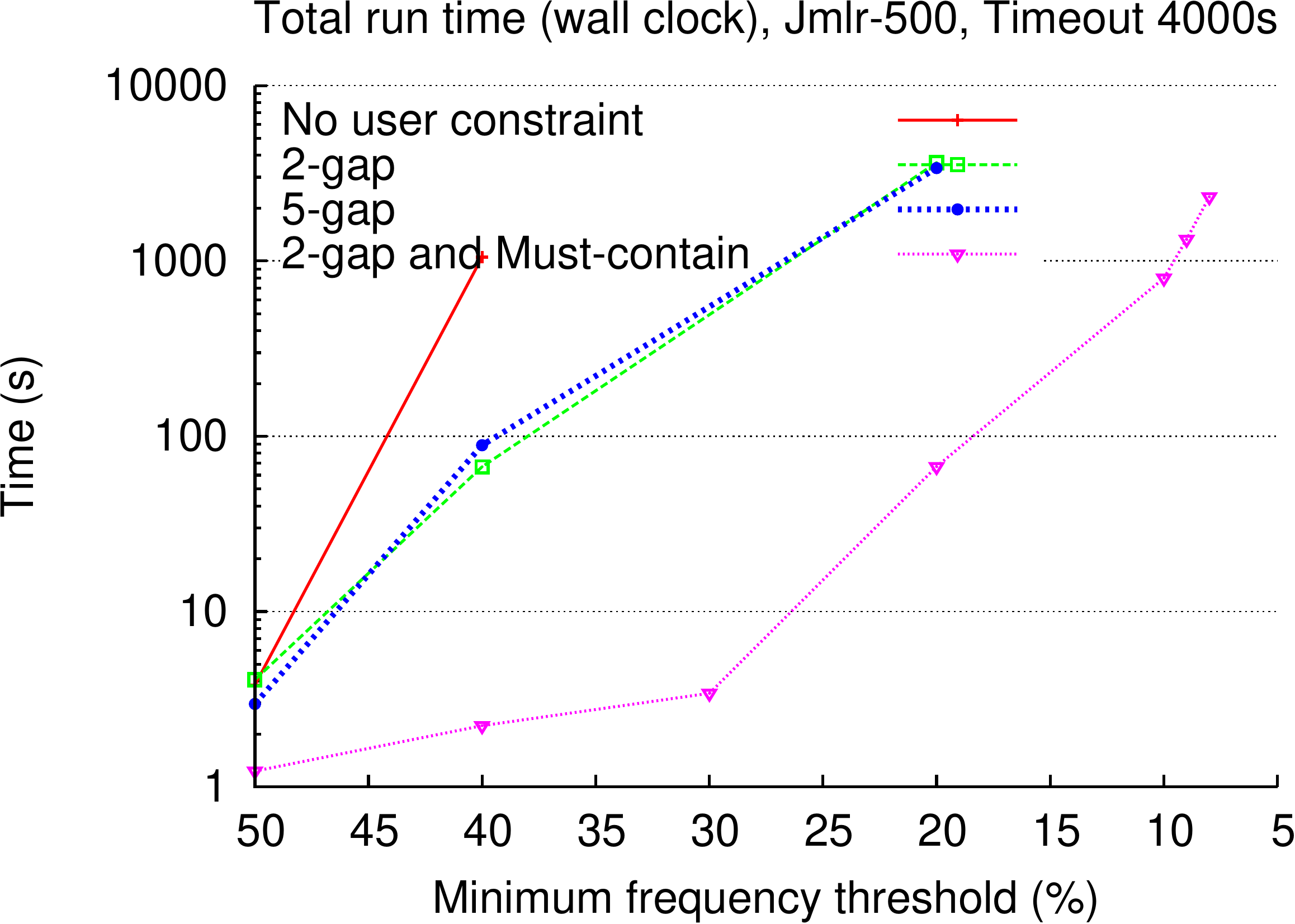}
      \caption{Impact of user constraints and total solving times on {\em JMLR-500}.} %Must-contain constraint is {\em must contain machine AND learning}.}
\label{fig:jmlr500a}
\end{figure}

\begin{figure}[h!]
 \centering
 \includegraphics[width=0.4\textwidth]{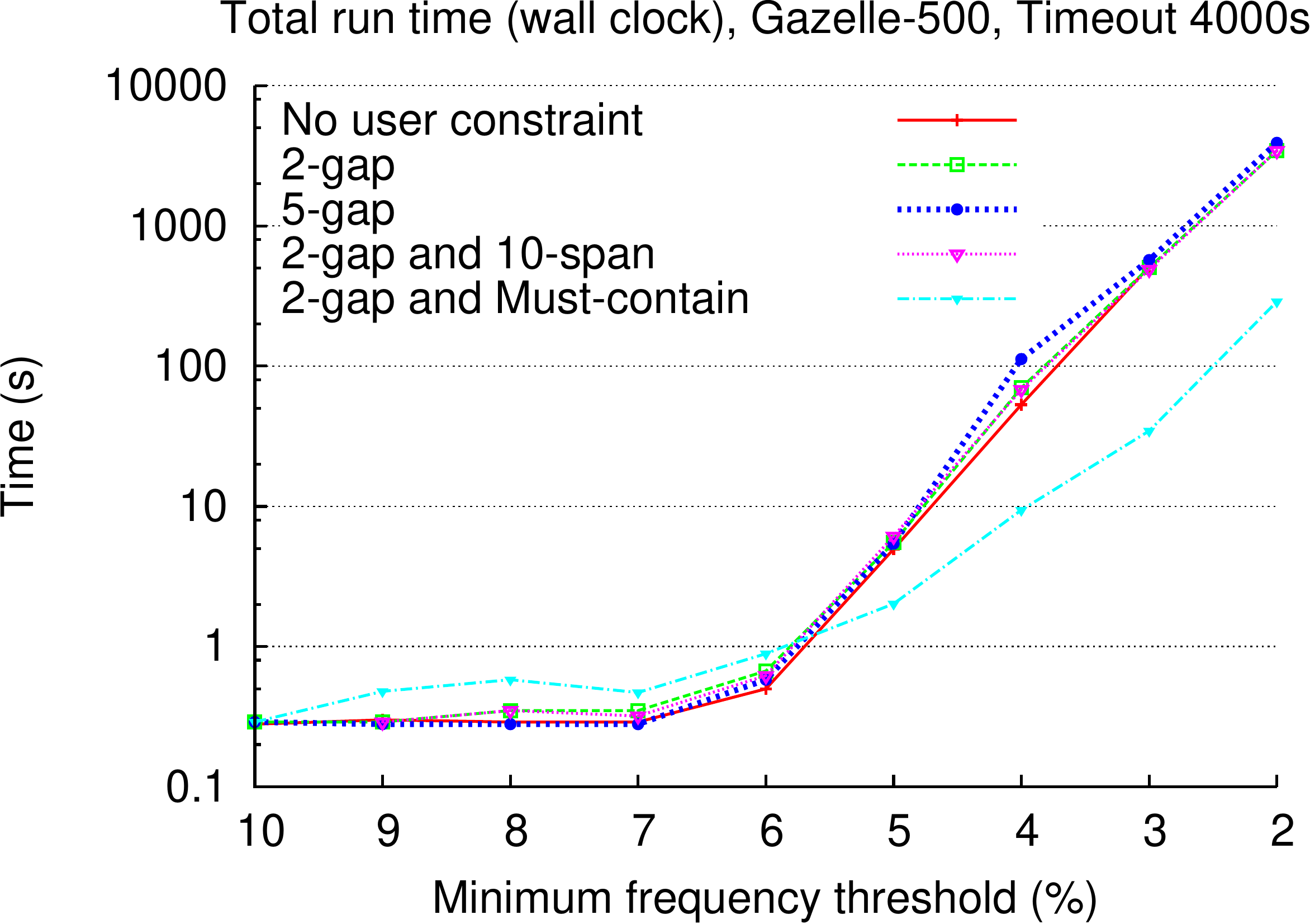}  
 \caption{Impact of user constraints and average solving times on {\em Gazelle-500}. Must-contain constraint is {\em must contain vocable-16}.}
  \label{fig:gaz500a}
\end{figure}

The total run times are presented in Figure~\ref{fig:jmlr500a}  for {\em
  JMLR-500} and Figure~\ref{fig:gaz500a}  for {\em Gazelle-500}.
  
\begin{figure}[h]
 \centering
 \includegraphics[width=0.4\textwidth]{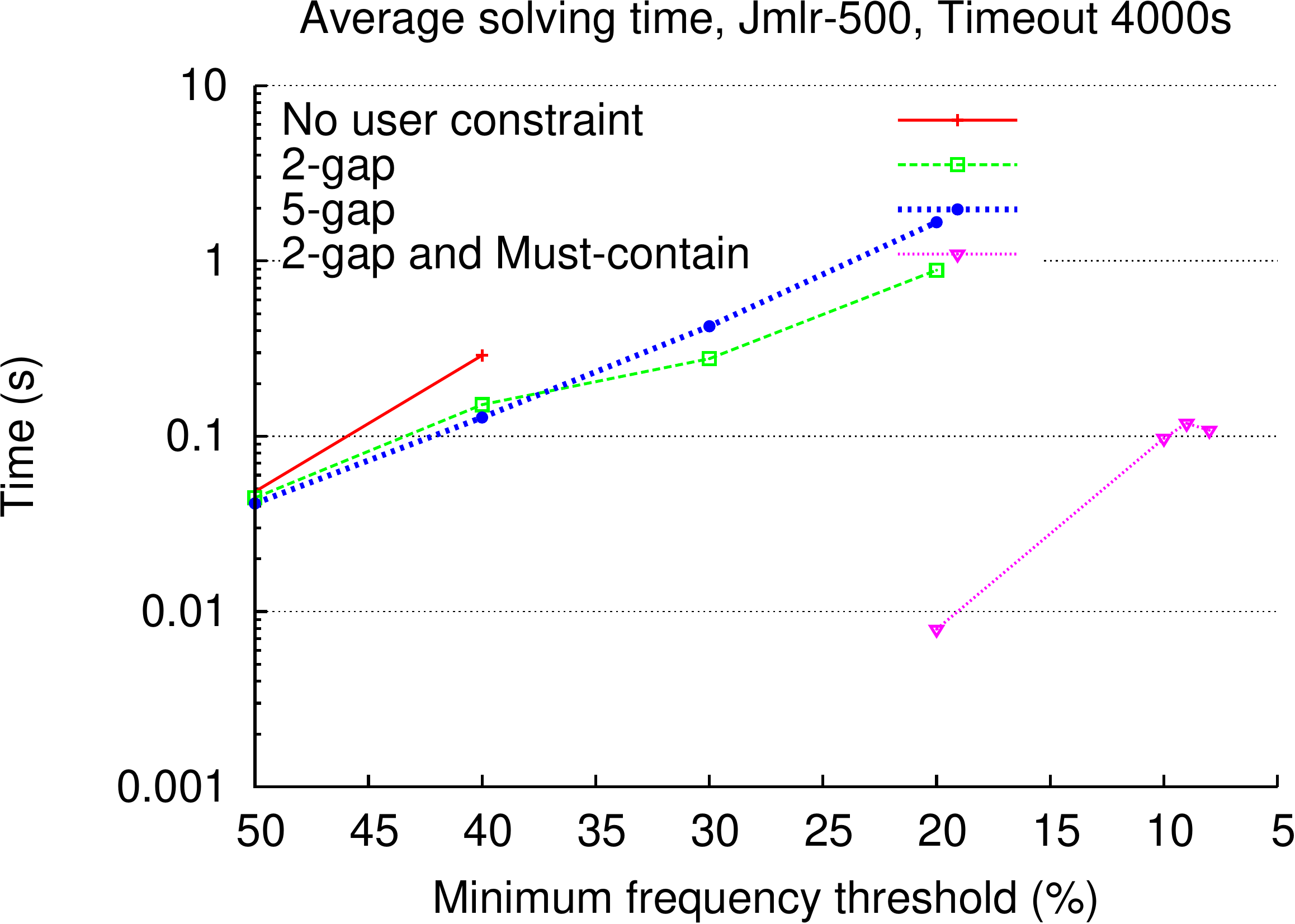} 
  \caption{Impact of user constraints and average solving times on {\em JMLR-500}.} %Must-contain constraint is {\em must contain machine AND learning}.}
 \label{fig:jmlr500b}
\end{figure}

\begin{figure}[h!]
 \centering
 \includegraphics[width=0.4\textwidth]{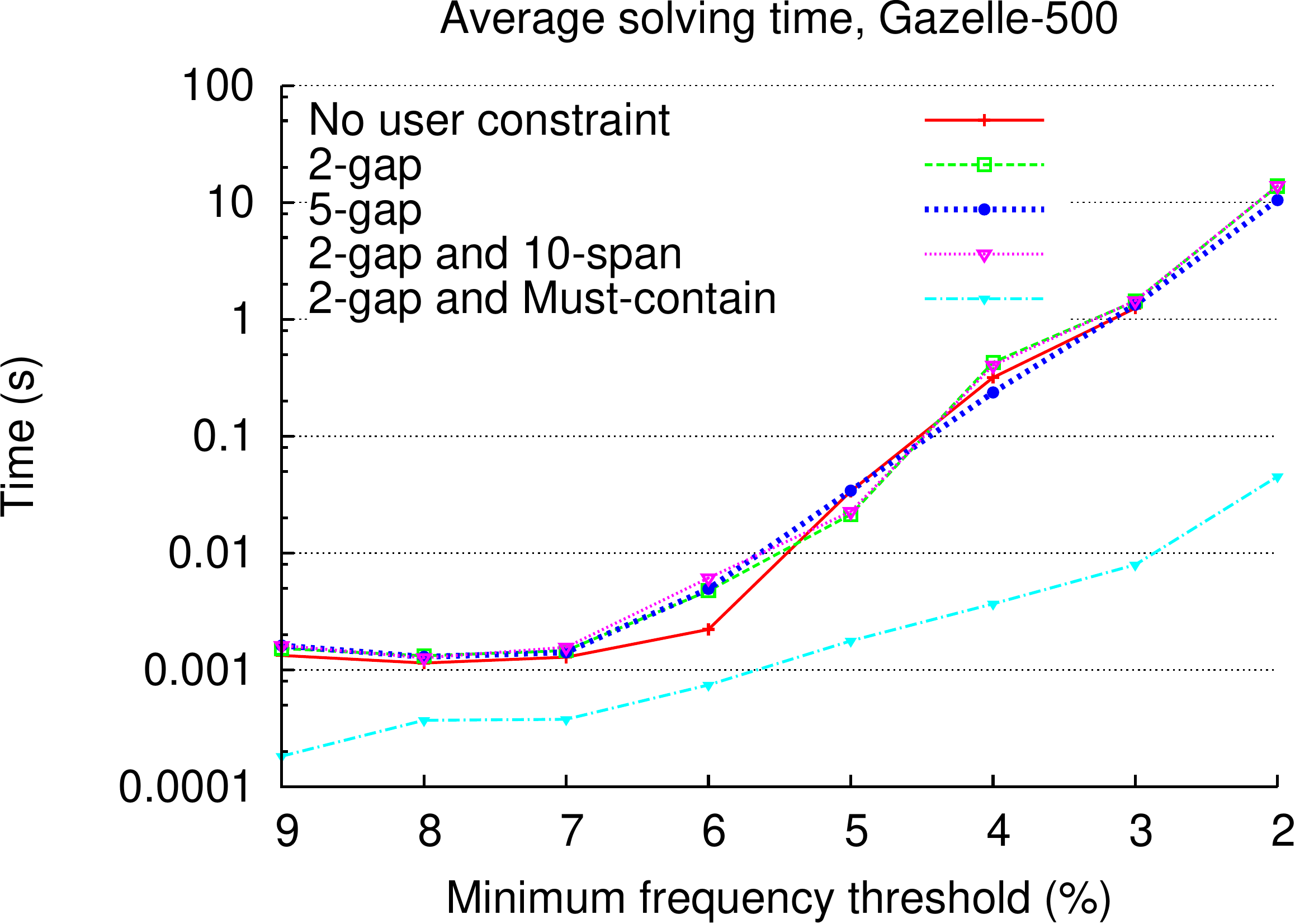} \caption{Impact of user constraints and average solving times on {\em Gazelle-500}. Must-contain constraint is {\em must contain vocable-16}.}
  \label{fig:gaz500b}
\end{figure}

It is important to note that, from the solver's point of view, adding
constraints harden the problem but also reduces the number of required %Syl: harden? A voir !
solver calls (Algorithm~\ref{algo:solveAll}, line~3). In order to
demonstrate that each individual call to the solver performs faster, %Syl than what?
we also measure the average solving time per run. The average solving
times are presented in Figure~\ref{fig:jmlr500b}  for {\em
  JMLR-500} and Figure~\ref{fig:gaz500b} for {\em Gazelle-500}.

% The two following experiments aims at evaluating of using a generic SAT
% solver over specialized algorithms. Specialized algorithm are very
% efficient but they are hard to modify address variations of the
% pattern mining task they were designed for. The experiments presented
% in Section~\ref{sec:sat-solver-vs} aim at comparing the raw efficiency
% of SAT versus specialized algorithms. The experiments presented in
% Section~\ref{sec:adding-extra-constr} aims at demonstrating that,
% given any extra user constraint, the SAT solver is able to prune the
% search space according to this new constraint. 

%   Currently the only
% options that is left to the pattern mining practitioner with a
% specific constraint is the run an existing problem and then
% post-process the generated patterns to extract the relevant patterns.
% Not only, the post-processing can take time, but the specialized
% algorithms have to explore a search space of patterns that is
% uselessly large.  In this experiment we demonstrate that the SAT based
% approach is slower but is able to reduced the search space and
% simplify the mining process as we add more user constraints. Put in a
% different way, the SAT solver is able to prune the search space in a
% way a specialized solver would do it.

We first observe that, on {\em JMLR-500}, the
gap constraint is very effective at reducing the total run time and the
average solving time. However, on {\em Gazelle-500}, the gap and the span %Syl: very little influence?
constraints have low impact.  Indeed, the vast
majority of sequences in  {\em Gazelle-500} are very small in size
(c.f. Table~\ref{tab:datasets}). As a consequence, gap and span
constraints do not reduce the number of solutions. Clearly, however,
the addition of extra constraints does not negatively impact
performance. 

The combination of the regular constraints with the 2-gap constraints is very effective, allowing %Syl: 2-gap
the solver to drastically reduce mining times. %remi : définir proprement maxgap(2)

%
%\begin{figure}[h!]
%%      \vspace{-0.5cm}
%  \begin{minipage}{0.5\linewidth}
%    \includegraphics[width=0.95\textwidth]{constraints_gaz}    
%  \end{minipage}\hfill%
%  \begin{minipage}{0.5\linewidth}
%    \includegraphics[width=0.95\textwidth]{solve_times_gaz}    
%  \end{minipage}%
%  \centering
%  \caption{Impact of user constraints and average solving times on {\em Gazelle-500}. Must-contain constraint is {\em must contain vocable-16}.}
%  \label{fig:gaz500}
%  %  \vspace{-0.6cm}
%\end{figure}

% \begin{figure}[h!]
%   \begin{minipage}{0.5\linewidth}
%     \includegraphics[width=0.95\textwidth]{constraints}    
%   \end{minipage}\hfill%
%   \begin{minipage}{0.5\linewidth}
%     \includegraphics[width=0.95\textwidth]{delay}    
%   \end{minipage}%
%   \centering
%   \caption{Impact of extra constraints}
%   \label{fig:constraints}
% \end{figure}

% L'idée générale est de montrer que sur les problèmes de base (sans contraintes additionnelles) 
% et avec un minsup très bas, on se fait éclater.

% Puis on durcit, en rajoutant toutes sortes de contraintes (sur les patterns sur les gaps, etc..)

% On arrive à la conclusion: 
% As shown in the experimental section of \cite{DeReadt08}, CP models outperform  dedicated data-mining
% algorithms, only when there are some extra-constraints over the pattern we look for.

% Tester aussi les différentes améliorations qu'on a proposé à la section \ref{sec:exploting},
% par exemple virer ou non les caractères infréquents, réordonner ou non les caractères par fréquence. 

%PS: penser à donner l'exemple sur lequel LCMSeq est buggé 

\section{Related work}
\label{sec:related-work}

% As pointed by \cite{lucsurvey}, there exist a broad variety of work
% about sequence mining. 

Research on sequence mining was initiated by Agrawal et
al.\cite{agrawal1995mining}, who proposed an Apriori-derived algorithm
to mine frequent sequences of itemsets (as opposed to mining frequent sequences
of single characters, which is the issue that we
address in this paper).  The problem of mining sequences of single
characters was introduced by Mannila et al. in
\cite{mannila1997discovery} as the problem of {\em serial episode
mining}. The authors proposed an Apriori-based algorithm with a depth-first search
strategy. 
However, depth-first search {\em level wise} pattern mining %Syl: vérifier l'ordre
algorithms usually require great memory capacity, because of the need
to store all the patterns of size $n$ before enumerating patterns of 
size $n+1$.  An important improvement was therefore proposed by Wang et al.
in \cite{wang2004bide} with the first depth-first search BIDE
algorithm.  In addition, these authors proposed a definition of closed
patterns for sequence mining. The techniques introduced by BIDE enabled
it to perform faster and to lower memory requirements by several orders of magnitude.%  More
% recently in \cite{OhtaniKUA09}, Arimura and Uno proposed another
% sequence mining algorithm, based on the notoriously efficient itemset
% mining algorithm LCM. % LCM\_seq introduce another definition of closed
% % sequence that they called {\em output extension closure} which
% performs better at reducing the number of redundant patterns.  This is
% the definition that we use in this paper.

Attempts to improve the flexibility of existing
sequence mining algorithms have been few. 
%For example the cSPADE algorithm
%(\cite{zaki2000sequence}) is an extension of SPADE for mining frequent
%sequences of itemsets, that can handle several extra user constraints
%such as constraint on the size of the pattern, gap constraints, window
%size constraint or other. 
In \cite{albert2003framework}, the authors mine sequences that satisfy a
regular expression. The LCMseq algorithm (\cite{OhtaniKUA09}) also deserves mention: 
It was extended by its authors to address different variations of the sequence mining task, such
as maximal sequences and sequences with gap constraints.  
Unfortunatelly, it was proven incorrect (Section \ref{sec:solv-deriv-sequ}).
%The main limitation of 
Moreover, the above approaches for mining
sequences with extra user constraints suffer from an important limitation: they can only support a
small set of built-in constraints, and they fail to propose a general framework for the addition of extra constraints.

A more flexible approach was proposed by Guns et
al. (\cite{DeReadt08}). In their work, these authors demonstrated that CP solvers and CP
modeling techniques can be used to efficiently address various types of itemset
mining problems. 
Since then,  data mining and SAT/CP experts have obtained
more results in the same vein. For example, Metivier et
al.  have proposed a constraint-based language for itemset mining
(\cite{metivier2011constraint}). More recently, in \cite{lakdar12},
Coquerry et al. have proposed to perform simple sequence mining tasks
using a SAT formulation.% of the problem. 
To the best of our knowledge, this
represents the first attempt to address the sequence mining problem with %Syl: inverser
the use of a SAT solver. However, this formulation is based on a set encoding of %Syl: set encoding
sequences proposed by Arimura and Uno (\cite{arimura2009polynomial})
and cannot be extended to more complex mining tasks such as flexible
sequences.

\section{Conclusion \& Perspectives}
\label{sec:conclu}
In this paper, we first proposed SAT encoding to adress the 
sequence mining problem. Second, we formulated several constraints, such as gap or span constraints,
which  are relevant to mining in various application fields. third, we proposed a sound methodology for addressing complex mining tasks, such as closed and maximal pattern
mining. Further, we implemented this methodology with the Glucose SAT
solver using interactive SAT solving. 

To date, while state-of-the-art sequence mining algorithms are
faster at generating all the frequent sequences, this approach is not %Syl: get rid of "all the"
scalable: since the number of sequences grows exponentially with the
size of the dataset, unconstrained mining of sequences is bound to
intractability.
%\todo{revoir cette conclu}
Our study and our experiments not only demonstrate the feasibility of
using SAT for solving sequence mining problems, but also highlight
several important benefits for data-mining users, including the ability to
combine any arbitrary constraints.%  where specialized algorithms can
% only handle a small subset of hard coded constraints at best, no

%Over the last decade, many optimizations have been proposed in the pattern
%mining community.  A natural extension of this work is to formulate
%these optimizations in our model, we already successfully integrate some of the 
%data-mining techniques techniques in our SAT model, 
%%One simple example that we have
%%mentioned in the paper, is the frequency based ordering of the
%%characters, 
%but there exist many other such as dataset reduction techniques
%(\cite{uno2004lcm}), or the exploitation of structural properties of
%the search space (\cite{arimura2009polynomial}) we plan to investigate.  
%Other work such as (\cite{negrevergneminer} 

In \cite{negrevergneminer}, the authors showed that highly
optimized pattern mining algorithms are inherently difficult to parallelize
due to the large number of instances of memory access. Using a SAT solver to tackle mining problems %Syl: memory accesses
 is a very different approach involving different algorithmic
behavior and memory access patterns. Studying how our SAT-based approach may resolve the memory access problem presents an interesting research challenge.

%\section{Acknowledgments}
%TODO commenter si soumission anonyme \\
%Thanks to G. Audemard and L. Simon for their help with Glucose source code, T. Guns and S. Nijssen for insightful discussions.
%This work was supported by the Research Foundation---Flanders under the project ``Principles of Patternset Mining'' and by the European Commission under the project ``Inductive Constraint Programming", contract number FP7-284715.
%
%

\bibliography{biblio}

\end{document}